\documentclass[10pt,twocolumn,letterpaper]{article}

\usepackage{cvpr}
\usepackage{times}
\usepackage{epsfig}
\usepackage{graphicx}
\usepackage{amsmath}
\usepackage{amssymb}
\usepackage{multirow}
\usepackage{array}
\usepackage{booktabs}

\newcommand{\bB}{\mathbf{B}}

\newcommand{\bI}{\mathbf{I}}

\newcommand{\bp}{\mathbf{p}}

\newcommand{\bT}{\mathbf{T}}

\newcommand{\bx}{\mathbf{x}}

\newcommand{\nR}{\mathbb{R}}

\newcommand{\figref}[1]{Fig.~\ref{#1}}

\newcommand{\eqnref}[1]{Eq.~\eqref{#1}}
\newcommand{\tabnref}[1]{Table~\ref{#1}}

\DeclareMathOperator*{\argmin}{argmin~}

\makeatletter
\DeclareRobustCommand\onedot{\futurelet\@let@token\@onedot}
\def\@onedot{\ifx\@let@token.\else.\null\fi\xspace}
\def\eg{e.g\onedot} 
\def\ie{i.e\onedot}

\def\etal{et~al\onedot}

\makeatother

\newcommand{\boldparagraph}[1]{\vspace{0.1cm}\noindent{\bf #1:} }

\newcommand{\norm}[1]{\left\lVert#1\right\rVert}

\usepackage[pagebackref=true,breaklinks=true,letterpaper=true,colorlinks,bookmarks=false]{hyperref}

\cvprfinalcopy % 
\begin{document}

%%%%%%%%% TITLE
\title{MBA-VO: Motion Blur Aware Visual Odometry}

\author{Peidong Liu$^{1}$\qquad Xingxing Zuo$^{1}$\qquad Viktor Larsson$^{1}$\qquad Marc Pollefeys$^{1,2}$ \\
	$^{1}$Computer Vision and Geometry Group, ETH Z\"{u}rich, Switzerland \\
	$^{2}$Microsoft Artificial Intelligence and Mixed Reality Lab, Z\"{u}rich, Switzerland \\
	{\tt\small \{peidong.liu, xingxing.zuo, vlarsson, marc.pollefeys\}@inf.ethz.ch}\\
	{\tt\small mapoll@microsoft.com}
}

\maketitle
\thispagestyle{empty}

%%%%%%%% Main text
%%%%%%%%% ABSTRACT
\begin{abstract}
Motion blur is one of the major challenges remaining for visual odometry methods. In low-light conditions where longer exposure times are necessary, motion blur can appear even for relatively slow camera motions. In this paper we present a novel hybrid visual odometry pipeline with direct approach that explicitly models and estimates the camera's local trajectory within the exposure time. This allows us to actively compensate for any motion blur that occurs due to the camera motion. In addition, we also contribute a novel benchmarking dataset for motion blur aware visual odometry. In experiments we show that by directly modeling the image formation process, we are able to improve robustness of the visual odometry, while keeping comparable accuracy as that for images without motion blur. 
\end{abstract}
\section{Introduction}
% background, motivation
Visual odometry (VO) determines the relative camera motion from captured images. As a fundamental block for many vision applications, such as robotics and virtual/augmented/mixed reality, great progress has been made during the last two decades. 
There have been many algorithms proposed in the literature: ranging from classical geometric approaches, deep learning based approaches to hybrid approaches. The geometric approaches recover the motion based on multi-view geometric constraints. Both the reprojection error (\eg ORB-SLAM~\cite{mur2017orb}) and the photometric consistency (e.g.~DSO~\cite{engel2017direct}) are commonly used constraints for the optimization. Deep learning based approaches formulate the problem as an end-to-end regression problem. Current state-of-the-art networks are still not able to achieve comparable performance to the classical approaches for large scale environments. Hybrid approaches usually embed a deep network inside a classical pipeline, to further improve their accuracy and robustness.

While many state-of-the-art algorithms have been proposed, motion blur is still a major challenge remaining for visual odometry methods. Motion blur is one of the most common artifacts that degrade images. It usually occurs in low-light conditions where longer exposure times are necessary. This affects both feature based approaches (e.g. ORB-SLAM~\cite{mur2017orb}), which struggle to detect keypoints, and direct methods (e.g.~DSO~\cite{engel2017direct}) which rely on strong image gradients for their alignment. While relocalization strategies can partially mitigate the problem by allowing the VO to recover after losing track, VO would still fail if the camera continues to move in un-explored areas.

%With the recent success on image deblurring by deep neural networks \cite{Nah2017CVPR, Tao2018CVPR, Kupyn2018CVPR, Kupyn2019ICCV}, one simple solution is to deblur every input image as a preprocessing step to standard visual odometry pipelines. However, we argue that it is not the correct approach due to two reasons:

%1) Although recent deep network achieves remarkable performance on image deblurring, they still struggle to recover high quality sharp images from severely motion blurred input. In experiments we show that these recovery artifacts can still degrade the performance of visual odometry and lead to tracking failures.
%There is still a chance that those poorly deblurred images would degrade the performance (or even lead to failure) of the VO.

%2) State-of-the-art image deblurring networks require significant computation effort and usually cannot run in real time even with a high-end GPU  \cite{Nah2017CVPR,Tao2018CVPR,Kupyn2018CVPR}. This excludes their use for most visual odometry application scenarios. Recently Kupyn et al.~\cite{Kupyn2019ICCV} proposed a more efficient network for deblurring. However, we experimentally found that it has limited generalization performance for VO applications due to its reduced model capacity.

\begin{figure}
    \centering
    \setlength\tabcolsep{0pt}
    \begin{tabular}{lr}
     	\includegraphics[width=0.5\columnwidth]{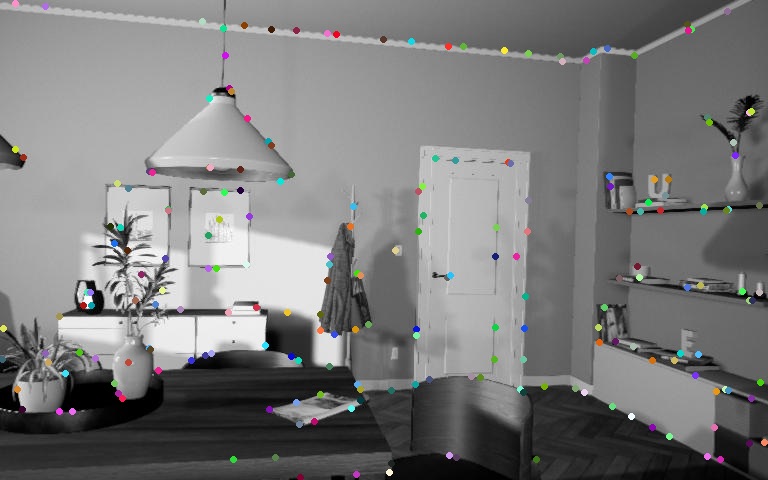}   	\includegraphics[width=0.5\columnwidth]{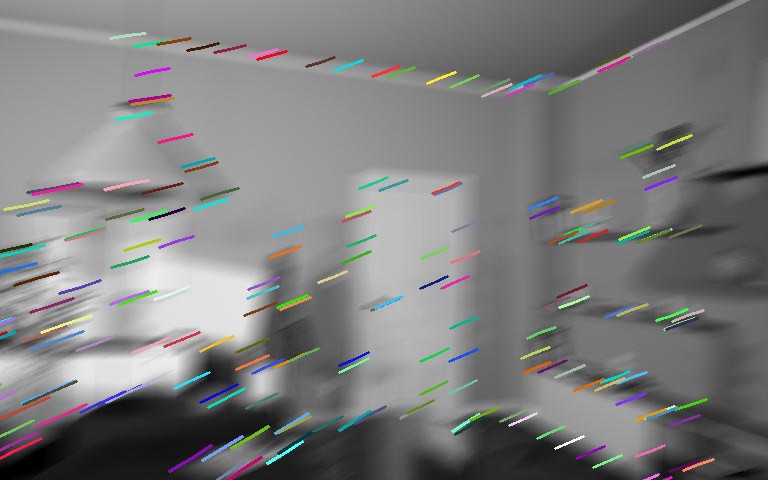}
    \end{tabular}
    
    \caption{\textit{Motion Blur Aware Visual Odometry}. We propose a full pipeline for performing motion blur aware visual odometry. By explicitly modelling the image formation process during tracking, we can actively compensate for motion blur in the direct image alignment.}
    \label{fig:my_label}
\end{figure}

In this paper, we thus propose a novel hybrid visual odometry method which is robust to motion blur. As conventional algorithms, our method consists of a front-end tracker and a back-end mapper. During tracking, instead of estimating the camera pose at a particular point in time, we estimate the local camera motion trajectory within the exposure time for each frame. This allows us to explicitly model the motion blur in the image and leverage it for tracking. We assume that the reference keyframe image is sharp, which is achieved by applying a deep deblurring network on the original motion blurred image. Since keyframes are usually sampled with a frequency much lower than frame-rate (and are less sensitive to latency), we can thus take advantage of a powerful deep network for keyframe deblurring (e.g.~\cite{Tao2018CVPR}). To estimate the camera motion trajectory during image capture, we locally re-blur the sharp reference keyframe image that is then used for direct image alignment against current tracked frame. 
%The estimated motion trajectory can be further used to measure the severity of the motion blur in current image, which serves as a guide for future keyframe selection. In particular, since we only select keyframes that have less motion blur initially, the problem becomes easier for the deep deblurring network which can then produce sharp results. 
The back-end jointly optimizes the camera poses and scene geometry based on the deblurred keyframe images, by maximizing the photometric consistency. We build our method on the popular DSO~\cite{engel2017direct} framework. 
% but we believe that our overall approach could be used as a drop-in replacement in many other VO systems, including feature-based approaches (such as ORB-SLAM~\cite{mur2017orb}). 

As another contribution, we also propose a novel benchmarking dataset targeting motion blur aware VO. Our dataset contains sequences with varying levels of motion blur. Time synchronized ground truth trajectories are also provided by an accurate indoor motion capturing system. By making this dataset publicly available to the community, we hope to encourage further research on making VO robust, which is important for real-world deployments.

We evaluate our approach with both synthetic dataset and real datasets. The experimental results demonstrate that we are able to improve the robustness of the visual odometry, while keeping comparable accuracy as that for images without motion blur. Furthermore, our motion blur aware VO (called \textit{MBA-VO}) is also able to run in real-time on a laptop with a Nvidia GeForce RTX 2080 graphic card. %We believe our algorithm and dataset would pave the way for the era of robust visual odometry.

%In this paper we propose a novel visual odometry method that estimates the local camera motion during the image capture for each frame (within exposure time). This allows us to explicitly model the motion blur in the image and leverage this during tracking. For each frame, we locally re-blur the reference key-frame image that is then used for direct alignment. We build our method on the popular DSO\cite{engel2017direct}, but we believe that our approach could be used as a drop-in replacement in many other VO systems, including feature-based approaches (such as ORB-SLAM\cite{mur2017orb}). In experiments we show that \todo{...}

\section{Related Work}

\boldparagraph{Visual odometry} Existing works on visual odometry can be categorized into three main groups: classical geometric approaches, deep learning based approaches and hybrid approaches. 

Classical geometry-based approaches recover the camera motion from multi-view constraints. These methods can be further divided into direct approaches and feature-based approaches. Direct approach relies on the photometric consistency assumption across multiple view within a short time interval. They jointly optimize the camera poses, 3D scene structure as well as camera intrinsic parameters by maximizing the photometric consistency. The representative works are LSD-SLAM \cite{Engel2014ECCV}, DSO \cite{engel2017direct} and their many variants \cite{schubert2019rolling, Liu2017IROS, Liu2018IROS, gao2018ldso}. Different from direct method, feature-based methods extract a set of sparse keypoints from the raw images which are then matched across different views. Both the camera poses and 3D scene geometry are estimated by enforcing consistency between the keypoint locations and the projections of the scene structure. %for the set of sparse keypoints.
Dating back from the early work by Davison et al.~\cite{Davison2003ICCV} and Nister et al.~\cite{Nister2004CVPR}, to the more recent ORB-SLAM \cite{mur2017orb}, many feature-based approaches have been proposed in the literature. More details can be found from a recent review paper from Cadena et al.~\cite{Cadena2016TOR}.
Deep learning-based approaches usually formulate the problem as an end-to-end regression problem. Although several pioneering works \cite{zhou2018deeptam, Zhou2017CVPR, Ummenhofer2017CVPR} and their variants have been proposed in the past years, they are still in their infancy compared to geometric approaches in terms of scalability and performance.
Recently there have been a series of hybrid methods~\cite{tateno2017cnn, bloesch2018codeslam, zhi2019scenecode}, which try to embed deep networks into classical geometric frameworks. These frameworks aim to leverage the benefits of both approaches to robustify the visual odometry.
%Hybrid approach tries to embed deep networks into classical geometric framework, 
%such that the original geometric methods can be improved  

Almost all those algorithms assume the input images are of good quality. However, due to environmental conditions (e.g.~low light), low quality of image is sometimes unavoidable in real world applications, which can then drastically reduce the performance of VO systems. In this paper we propose to tackle one of the most common challenging cases, motion blurred images.
The early works from Pretto \etal \cite{pretto2009visual} and Lee \etal \cite{lee2011simultaneous} have been proposed to improve the robustness of sparse keypoint based VO against motion blur. Pretto \etal \cite{pretto2009visual} propose to detect motion blur robust sparse invariant features. The work from Lee \etal \cite{lee2011simultaneous} is perhaps the one which is most similar to ours. In~\cite{lee2011simultaneous}, the authors assume the motion between neighbouring frames is smooth and try to linearly interpolate the motion within the exposure time. For each frame the initial motion is extrapolated from previous frames using a motion model and this prediction is used to re-blur the patches from the keyframe. The re-blurred patches are used to establish explicit sparse correspondences between the new frame and the keyframe. The camera poses and scene structure are then estimated from these correspondences.
In this work we take a similar approach to \cite{lee2011simultaneous}, where we re-blur patches extracted from the keyframe. In contrast to~\cite{lee2011simultaneous}, which relies heavily on the initial motion prediction to make hard decisions on correspondences, we directly optimize the local camera motion trajectory used to re-blur the patches and instead implicitly solve the data association problem using a direct image alignment approach.

\boldparagraph{Image deblurring}
Motion deblurring methods can be categorized into classic optimization based approaches and modern deep learning based approaches. We will only focus our attention on several representative single image deep deblurring networks, since they are most related to our work. 
Recently, the performance of single image deblurring algorithms has been boosted significantly by deep neural networks. 
The early work for image deblurring by Xu \etal~\cite{Xu2014NIPS} is a shallow network with four hidden layers, which is trained end-to-end with known ground truth sharp images. Hradis \etal \cite{Hradis2015BMVC} later propose a 15 layer network for text image deblurring. The network was further enlarged to 40 layers in a multi-scale manner by Nah \etal \cite{Nah2017CVPR}, resulting a network with 120 layers for three pyramid scales. Adversarial loss \cite{Goodfellow2014nips} was also introduced to improve the deblurring performance by Kupyn \etal \cite{Kupyn2018CVPR}. Another two concurrent works from Tao \etal \cite{Tao2018CVPR} and Zhang \etal \cite{Zhang2018CVPR} also achieve the state-of-the-art performance by using recurrent neural networks (RNN). Although these networks achieve remarkable performance, they usually cannot run at frame-rate even with a high-end GPU. Recently, Kupyn \etal \cite{Kupyn2019ICCV} propose a light-weight network which is able to run in real-time, with an expense of slightly downgraded deblurring quality. In this work, we will explore the networks from Tao \etal (better quality but slow) \cite{Tao2018CVPR} and Kupyn \etal (lower quality but fast) \cite{Kupyn2019ICCV}.

\boldparagraph{Existing dataset for robust visual odometry}
In the last decade there have been several benchmark datasets proposed for evaluating visual odometry and SLAM methods \cite{Sturm2012IROS,geiger2013vision,handa2014benchmark,blanco2014malaga,burri2016euroc,pfrommer2017penncosyvio,majdik2017zurich,Schops2019CVPR}. Some datasets focus on evaluating specific aspects or settings; e.g.~autonomous-driving~\cite{geiger2013vision}, illumination changes~\cite{park2017icra} and long-term relocalization~\cite{carlevaris2016university}.
In this paper we propose a new benchmark dataset for evaluating visual odometry which specifically targets at motion blur. While images with motion blur appear in some of the previous datasets~(e.g.~\cite{Sturm2012IROS,Schops2019CVPR}, see Section~\ref{sec:datasets}), it is not their main focus. We provide sequences with varying degrees of motion blur, which allows us to more precisely evaluate the breaking points of different methods.

%KITTI~\cite{geiger2013vision}, TUM RGB-D~\cite{sturm12iros}, Malaga~\cite{blanco2014malaga}, EuRoC MAV \cite{burri2016euroc}, PennCOSYVIO \cite{pfrommer2017penncosyvio}, Zurich Urban MAV~\cite{majdik2017zurich}, TUM VI \cite{schubert2018tum}, ETH3D~\cite{schops2019bad} and 4Seasons \cite{wenzel20204seasons}.

\section{Method}
% overview of our pipeline
In this section we present our motion blur aware visual odometry. We build on Direct Sparse Odometry (DSO) from Engel et al.~\cite{engel2017direct}. The proposed pipeline consists of three main parts: a motion blur aware visual tracker, a keyframe deblurring network and a local mapper. 

The front-end tracker estimates the camera motion trajectory within the exposure time of current blurry frame, relative to the latest sharp keyframe image. Each new keyframe is processed with the motion deblurring network. The local mapper then jointly optimizes the camera poses and the scene structure based on the recovered latent sharp keyframe images. We use the same local mapper as in DSO~\cite{engel2017direct} and the main technical contribution of our work is the motion blur aware tracker, which we will detail in the following sections.

\boldparagraph{Motion blur image formation model} The physical image formation process of a digital camera, is to collect photons during the exposure time and convert them into measureable electric charges. This process can be mathematically modelled as integrating over a set of virtual sharp images: 
\begin{equation}
\bB(\bx) = \lambda \int_{0}^{\tau} \bI_\mathrm{t}(\bx) \mathrm{dt},
\end{equation}
where $\bB(\bx) \in \nR^{\mathrm{W} \times \mathrm{H} \times 3}$ is the captured image, $\mathrm{W}$ and $\mathrm{H}$ are the width and height of the image respectively, $\bx \in \nR^2$ represents the pixel location, $\lambda$ is a normalization factor, $\tau$ is the camera exposure time, $\bI_\mathrm{t}(\bx) \in \nR^{\mathrm{W} \times \mathrm{H} \times 3}$ is the virtual sharp image captured at timestamp $\mathrm{t}$ within the exposure time. Motion in the camera during the exposure time will result in different virtual images $\bI_\mathrm{t}(\bx)$ for each $t$, resulting in a blurred image $\bB(\bx)$. The model can be discretely approximated as 
\begin{equation}\label{eq_blur_im_formation}
\bB(\bx)  \approx \frac{1}{n} \sum_{i=0}^{n-1} \bI_\mathrm{i}(\bx), 
\end{equation}
where $n$ is the number of discrete samples. 

The amount of motion blur in an image thus depends on the motion during the exposure time. For shorter exposure time, the relative motion will be small even for a quickly moving camera. Conversely, for long exposure time (e.g.~in low light conditions), even a slowly moving camera can result in a motion blurred image.

\boldparagraph{Direct image alignment with sharp images} Before introducing our direct image alignment algorithm with blurry images, we first review the original algorithm with sharp images. Direct image alignment algorithm serves as the core block for direct visual odometry approaches.
It jointly optimizes camera poses, scene structure as well as the camera intrinsic parameters by maximizing the photometric consistency across multiple images. For simplicity, we only consider optimizing over the relative camera pose here, but the approach extends naturally to the full problem. 
It can be formally defined as follows:
\begin{equation}\label{eq_photoconsistency}
\bT^{*} = \argmin_{\bT} \sum_{i=0}^{m-1} \norm{\bI_{\mathrm{ref}}(\bx_i) - \bI_{\mathrm{cur}}(\hat{\bx}_i)}_2^2,
\end{equation}
where $\mathrm{\bT} \in \mathbf{SE}(3)$ is the transformation matrix from the reference image $\bI_{\mathrm{ref}}$ to the current image $\bI_{\mathrm{cur}}$, $m$ is the number of sampled pixels for motion estimation, $\mathrm{\bx}_i \in \nR^2$ is the location of the $i^{th}$ pixel, $\hat{\bx}_i \in \nR^2$ is the pixel location corresponding to pixel $\mathrm{\bx}_i$ in current image $\bI_{\mathrm{cur}}$. Robust loss function (\eg huber loss) is usually also applied to the error residuals for robust pose estimation. The image points $\mathrm{\bx}_i$ and $\hat{\bx}_i$ are related by the camera pose $\mathrm{\bT}$ and the depth $d_i$ as
\begin{equation} \label{eq_transfer}
\hat{\bx}_i = \mathrm{\pi}(\bT \cdot \pi^{-1}(\bx_i, d_i)),
\end{equation}
where $\pi: \nR^3 \rightarrow \nR^2$ is the camera projection function, which projects point in 3D space to image plane; $\pi^{-1}: \nR^2 \times \nR \rightarrow \nR^3$ is the inverse projection function, which transforms a 2D point from image to 3D space by backprojecting with the depth $d_i$. 

Direct VO methods assume that photoconsistency (i.e.~equation~\ref{eq_photoconsistency}) holds for the correct transformation $\bT$. However, if the images $\bI_{\mathrm{ref}}$ and $\bI_{\mathrm{cur}}$ are affected by different motion blur, the photoconsistency loss will no longer be valid since the local appearance for correctly corresponding points will differ. This scenario is unavoidable in settings with highly non-linear trajectories, e.g.~tracking in augmented/mixed/virtual reality applications, which usually result in images with different levels of motion blur. 

\boldparagraph{Motion trajectory modeling} 
To correctly compensate for the motion blur we need to model the local camera trajectory during the exposure time. One approach is to only parameterize the final camera pose and then linearly interpolate between the previous frame and the new estimate. From the interpolation we can then create the virtual images necessary to represent the motion blur, as in equation~\eqref{eq_blur_im_formation}. However, this approach might fail for camera trajectories with very abrupt directional changes, which are quite common for hand-held and head-mounted cameras.

To ensure robustness, instead, we choose to parameterize the local camera trajectory independently of the previous frame.
To be specific, we parameterize two camera poses, one at the beginning of the exposure $\bT_\mathrm{start} \in \mathbf{SE}(3)$ and one at the end $\bT_\mathrm{end} \in \mathbf{SE}(3)$. Between the two poses we linearly interpolate poses in the Lie-algebra of $\mathbf{SE}(3)$. The virtual camera pose at time $t \in [0,\tau]$ can thus be represented as 
\begin{equation} \label{eq_trajectory}
\bT_t = \bT_\mathrm{start} \cdot exp(\frac{t}{\tau} \cdot log(\bT_\mathrm{start}^{-1} \cdot \bT_\mathrm{end})),
\end{equation} 
where $\tau$ is the exposure time. For more details on the interpolation and derivations of the related Jacobian, please see the supplementary material.

The goal of our motion blur-aware tracker is now to estimate both $\bT_\mathrm{start}$ and $\bT_\mathrm{end}$ for each frame. If the two poses are close, we know that the corresponding frame has very little motion blur. In this work we only considered linear interpolation between the two poses, but e.g.~higher order splines could be used as well which could then represent more complex camera motions. However, in our experiments we found that the linear model worked well enough, since the exposure time is usually relatively short.

\boldparagraph{Direct image alignment with blurry images} Our motion blur-aware tracker works by performing direct alignment between the keyframe, which we assume is sharp, and the current frame which can suffer from motion blur. To leverage photometric consistency in the alignment, we thus need to either de-blur the new frame or re-blur the keyframe. In our work we chose the latter since re-blurring is in general easier and more robust compared to motion deblurring, especially for severe motion blurred images. 

\begin{figure}
	\centering
	\begin{tabular}{cc}
		\includegraphics[width=0.45\columnwidth]{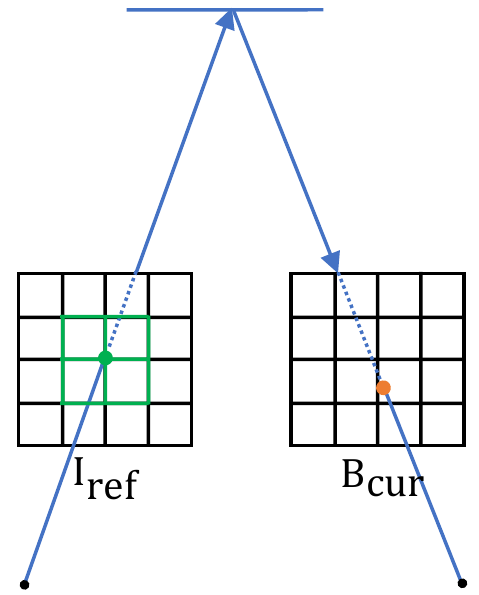} &
		\includegraphics[width=0.45\columnwidth]{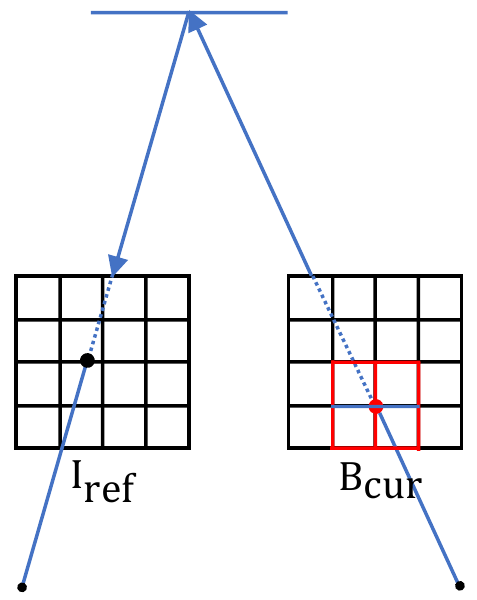} \\
	\end{tabular}
    \caption{Pixel point transfer strategies}
    \label{fig_image_warping}
\end{figure}

Each sampled pixel in $\bI_\mathrm{ref}$ with known depth can be transferred into the current (blurry) image $\bB_\mathrm{cur}$ using~\eqref{eq_transfer}. For each projected point we select its nearest neighbour integer position pixel, in current blurry image. Assuming that the 3D point lies on a fronto-parallel plane (with respect to $\bI_\mathrm{ref}$), we can use this plane to transfer the selected pixel back into the reference view. Details can be found in \figref{fig_image_warping}. To synthesize the re-blurred pixel from the reference view (so that we can compare against the real captured pixel intensity), we now interpolate between $\bT_\mathrm{start}$ and $\bT_\mathrm{end}$. For each virtual view $\bT_t$, which is uniformly sampled within $[0,\tau]$, we transfer the pixel coordinate (\ie the red pixel in \figref{fig_image_warping}) back into the reference image and retrieve the image intensity values using bi-linear interpolation. 
The re-blurred pixel intensity is then created by averaging over the intensity values (as in~\eqref{eq_blur_im_formation}):
\begin{equation}\label{eq_point_transfer}
\hat{\bB}_\mathrm{cur}(\bx) = \frac{1}{n}\sum_{i=0}^{n-1} \bI_\mathrm{ref}(\bx_{\frac{i\tau}{n-1}}),
\end{equation}
where $\bx_{\frac{i\tau}{n-1}} \in \nR^2$ corresponds to the transferred point at time $t = \frac{i\tau}{n-1}$ in the sharp reference frame.
The tracker then optimizes over the start-pose and end-pose to minimize the photoconsistency loss between the real captured intensities in current frame and the synthesized pixel intensities from the reference image (\ie via re-blurring),
\begin{equation}\label{eq_photoconsistency2}
\small
{\bT_\mathrm{start}^*,~\bT_\mathrm{end}^*} = \argmin_{\bT_\mathrm{start},~\bT_\mathrm{end}}  \sum_{i=0}^{m-1} \norm{\bB_\mathrm{cur}(\bx_i) - \hat{\bB}_{\mathrm{cur}}(\bx_i)}_2^2.
\end{equation}

In practice, most direct image alignment methods use local patches for better convergence. Different from direct image alignment algorithm for sharp images, which usually selects the local patch from the reference image (\eg the green $3\times3$ grid on the left of \figref{fig_image_warping}), we instead select the local patch from the current blurry image (\eg the red $3\times3$ grid on the right of \figref{fig_image_warping}) since this simplifies the re-blurring step of our pipeline. 

\boldparagraph{More details on the transfer} 
To further demonstrate the relationship between $\bx \in \nR^2$ and $\bx_{\frac{i\tau}{n-1}} \in \nR^2$ from \eqnref{eq_point_transfer}, we define following notations for the ease of illustration. We denote the depth of the fronto-parallel plane as $d$, which is the estimated depth of the corresponding sampled keypoint from $\bI_\mathrm{ref}$ (\ie the green pixel in \figref{fig_image_warping}); we further denote the camera pose of the virtual frame $\bI_i$ captured at timestamp $\frac{i\tau}{n-1}$ relative to the reference keyframe $\bI_{\mathrm{ref}}$ as $\bT_i \in \mathbf{SE}(3)$, which can be computed from \eqnref{eq_trajectory} as 
\begin{equation}
\bT_i = \bT_\mathrm{start} \cdot exp(\frac{i}{n-1}\tau \cdot log(\bT_\mathrm{start}^{-1} \cdot \bT_\mathrm{end})),
\end{equation}
where $\bT_\mathrm{start} \in \mathbf{SE}(3)$ and $\bT_\mathrm{end} \in \mathbf{SE}(3)$ are the camera poses (which are defined from the camera coordinate frame to the global world coordinate frame) of the current blurry image, at the beginning and end of the image capturing respectively, $\tau$ is the camera exposure time. 
\begin{figure}
	\centering
	\includegraphics[width=0.5\columnwidth]{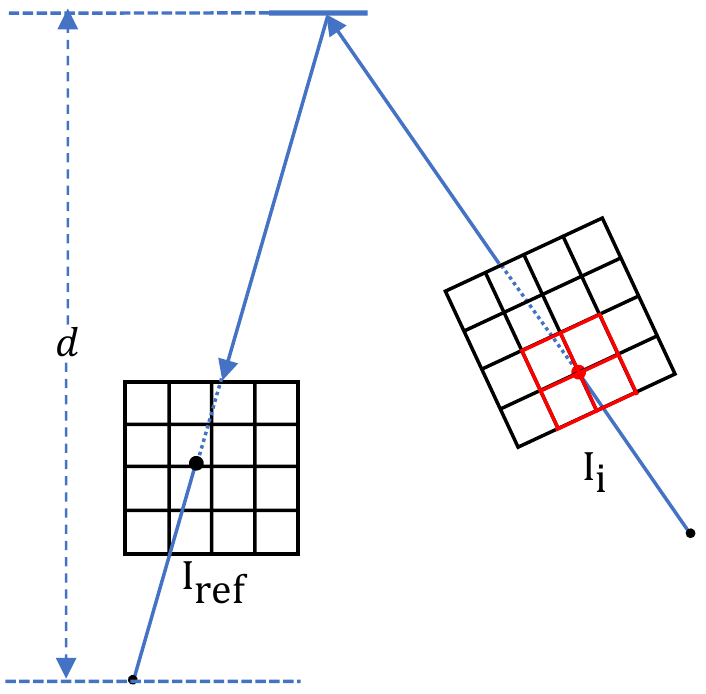}
	\caption{Geometric relationship between $\bx \in \nR^2$ (\ie the red pixel) of the virtual sharp image $\bI_i$ and $\bx_{\frac{i\tau}{n-1}} \in \nR^2$ (\ie the black pixel) of the reference image $\bI_\mathrm{ref}$.}
	\label{fig_geom_Ii}
\end{figure}
Note that the fronto-parallel plane is defined in the reference camera frame, it might not be fronto-parallel with respect to the $i^{th}$ virtual camera frame. To avoid confusion, we illustrate the relationship in \figref{fig_geom_Ii}. By proper algebraic manipulations, we can obtain $\bx_{\frac{i\tau}{n-1}}$ as 
\begin{align}
&\bx_{\frac{i\tau}{n-1}} = \pi (\bT_i \cdot \bp_\mathrm{3d}), \\
&\bp_\mathrm{3d} = \frac{d-p_z}{\lambda}
                   \begin{bmatrix}
                     x, &
                     y, &
                     z
                   \end{bmatrix}^\mathrm{T}, \\
&\lambda = 2x \cdot q_0 + 2y \cdot q_1 + z \cdot q_2 \\
&q_0 = q_x q_z - q_w q_y,\\
&q_1 = q_x q_w + q_y q_z,\\
&q_2 = q_w^2 - q_x^2 - q_y^2 + q_z^2, \\
&\begin{bmatrix}
  x, & y, & z
 \end{bmatrix}^\mathrm{T} = \pi^{-1}(\bx),
\end{align}
where $\pi: \nR^3 \rightarrow \nR^2$ is the camera projection function, $(q_w, q_x, q_y, q_z)$ is the quaternion representation of the rotation matrix of $\bT_i$ and $(p_x,p_y,p_z)$ is the translation vector of $\bT_i$, $d$ is the depth of the plane with respect to the reference key-frame, $\pi^{-1}: \nR^2 \rightarrow \nR^3$ is the camera back projection function such that $x^2 + y^2 + z^2 = 1$. Detailed algebraic derivations as well as related Jacobian can be found in our supplementary material.

\section{Datasets} \label{sec:datasets}
In this section we give an overview of the datasets that we consider in our experimental evaluation. While there are many different datasets for evaluating visual odometry methods, we found that there is no suitable dataset that specifically targets at motion blurred images, although some datasets have sub-sequences that contain motion blur, e.g.~in the ETH3D SLAM Benchmark~\cite{Schops2019CVPR} and TUM RGB-D~\cite{Sturm2012IROS}.

\boldparagraph{ETH3D~\cite{Schops2019CVPR} / ArchVizInterior}
In the ETH3D SLAM benchmark \cite{Schops2019CVPR}, the image sequences; \textit{camera\_shake\_1}, \textit{camera\_shake\_2} and \textit{camera\_shake\_3} have severe motion blur. The three sequences were captured with a camera being quickly shaken back and forth. In addition to the motion blur, the sequences are difficult due to the very poorly textured scene (mainly containing a white circular table with very few distinguishing landmarks). We experiment with both DSO \cite{engel2017direct} and ORBSLAM \cite{mur2017orb} on these sequences and find that both methods fail to initialize on this dataset.

\begin{figure}
	\centering
%	\begin{tabular}{cc}
%	\includegraphics[width=\columnwidth]{./gfx/4090_1061.jpg}
	\includegraphics[width=\columnwidth]{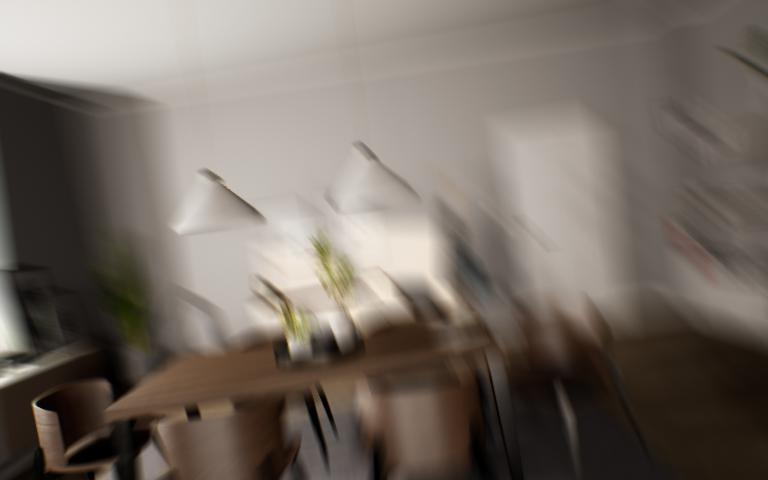}
%	\end{tabular}
	\caption{\textit{Sample image from our synthetic ArchVizInterior dataset}. The camera motions are taken from the ETH3D Benchmark~\cite{Schops2019CVPR} which are then re-rendered in synthetic scene by Unreal game engine.}
	\label{fig_archviz_sample}
\end{figure}

To investigate if the failures are due to the poorly textured scene or the motion blur, we render a synthetic photo-realistic dataset using the same motion trajectories, exposure time and frame rate. The dataset is rendered by the Unreal game engine with the free ArchVizInterior scene model\footnote{https://www.unrealengine.com}. We use the scripts provided by Liu \etal \cite{Liu2020CVPR} for the dataset creation. A sample image can be found in \figref{fig_archviz_sample}. Sample videos on the dataset can be found in our supplementary material. Since this dataset provides perfect ground truth sharp images, which are paired with the motion blurred images, we use it for our ablation studies.

\boldparagraph{TUM-RGBD~\cite{Sturm2012IROS}}
The hand-held SLAM sequences from the TUM-RGBD dataset \cite{Sturm2012IROS} also contain motion blurred images. The dataset is collected with the Microsoft Xbox Kinect sensor, which contains a rolling shutter color camera and a time-of-flight depth camera. Since the dataset targets at evaluating the performance of RGBD camera based SLAM methods, the effect of motion blurred images is not their focus. Furthermore, it has been shown that direct approach is more sensitive to rolling shutter effect \cite{Schops2019CVPR,Yang2018RAL}.
%by Sch\"ops \etal \cite{Schops2019CVPR}. 
It is thus better to have dataset which is collected from a global shutter camera to avoid the effect of rolling shutter mechanism. We also evaluate our method with the TUM-RGBD dataset in the next section. 

\boldparagraph{Proposed Motion Blur Benchmarking Dataset}
To more clearly show the benefit of our method we propose a new benchmark dataset for evaluating visual odometry which specifically targets motion blur. By making this dataset publicly available to other researchers, we hope to encourage further research on making visual odometry robust.

Our dataset is collected with a global shutter camera at a resolution of $752\times480$ pixels and a frame rate of 27 fps. The ground truth trajectory is provided by an indoor motion capturing system\footnote{https://www.vicon.com} at 100 Hz. Extrinsic parameters between the marker for motion capture and camera is calibrated by the hand-eye calibration approach, such that the ground truth trajectory can align with the camera motion trajectory. A total number of 18 motion blurred sequences are collected. The dataset contains images with varying levels of motion blur. More details on the dataset can be found in the supplementary video. Figure~\ref{fig:example_images} shows some example images from the new dataset with varying motion blur.

\begin{figure*}
    \centering
    \includegraphics[width=0.32\textwidth]{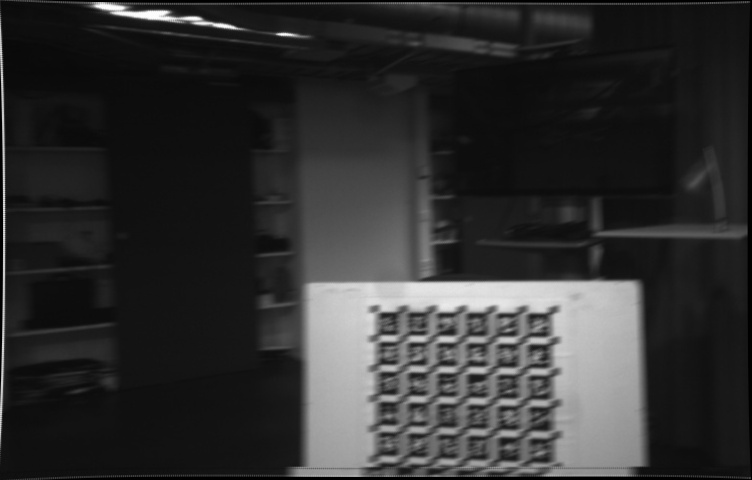}
    \includegraphics[width=0.32\textwidth]{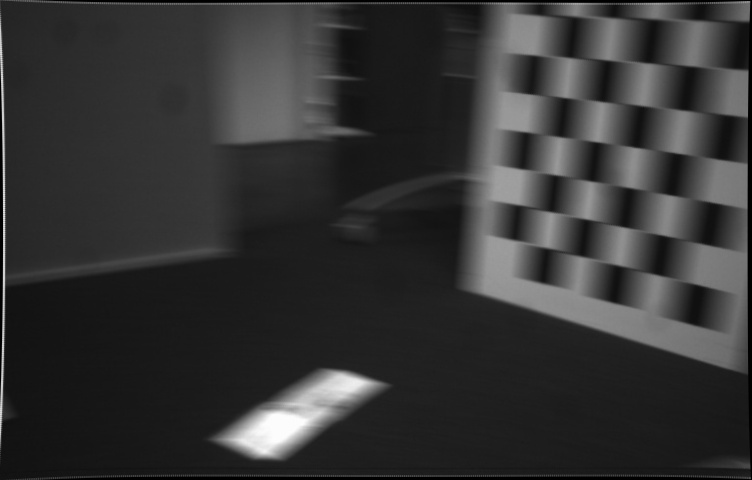}
    \includegraphics[width=0.32\textwidth]{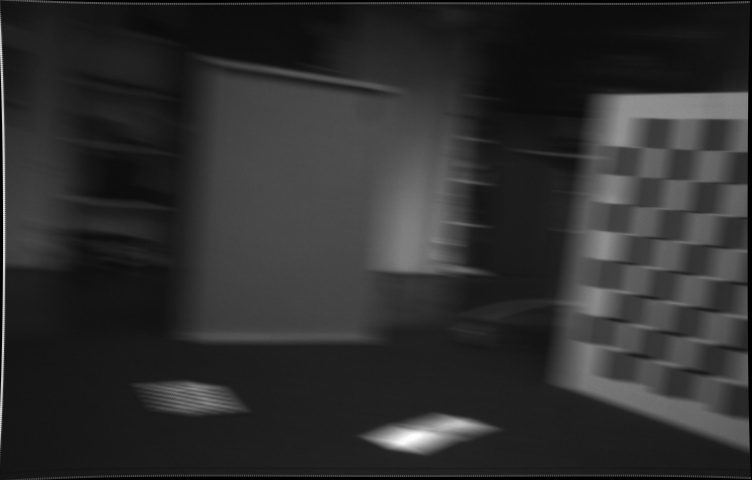}
    \caption{Examples images from the proposed dataset for benchmarking visual odometry from motion blurred image sequences. The dataset contains multiple sequences with varying levels of motion blur.}
    \label{fig:example_images}
\end{figure*}
\section{Experimental Evaluation}
\boldparagraph{Implementation details} The original tracker of DSO \cite{engel2017direct} does semi-dense direct image alignment. For efficiency, we sub-sample the high gradient pixels to obtain sparse keypoints, which are uniformly distributed within the image. We further define a $9\times9$ local patch around each sampled sparse keypoints for better convergence. The energy function is optimized in a coarse to fine manner and robust huber loss function is also applied for robustness. Our tracker is implemented and evaluated on a laptop grade Nvidia RTX 2080 graphic card. It takes 34.4 ms on average to process a single blurry image, which is suitable for real time applications.

We experiment with two state-of-the-art deblurring networks, SRNDeblurNet~\cite{Tao2018CVPR} and DeblurGANv2~\cite{Kupyn2019ICCV}. We use the official pretrained models and generalize them to our datasets without any finetuning. In particular, we consider the mobile network of DeblurGANv2 since it can run in real time on a high-end GPU. SRNDeblurNet takes around 140 ms second to process a single image at a resolution of $752\times480$ pixels on a laptop grade Nvidia RTX 2080 graphic card. However, it delivers higher quality deblurred images compared to the DeblurGANv2 mobile network and the time consumption is already sufficient for local mapping. There are also more advanced deblurring networks being proposed recently, such as the work from Gao \etal~\cite{Gao2019CVPR}. However, they are usually more time consuming than SRNDeblurNet~\cite{Tao2018CVPR}, and is not suitable to be integrated into our local mapper.

\boldparagraph{Baseline methods and evaluation metrics} Two state-of-the-art monocular VO pipelines are selected for the benchmark. In particular, we select ORB-SLAM \cite{mur2017orb}, which is the representative of sparse feature based approach. As a representative for direct approaches we compare with DSO~\cite{engel2017direct}. For quantative comparisons, we measure the RMSE of absolute trajectory error (\ie RMSE ATE) \cite{Sturm2012IROS}, since it is the focus of VO algorithms and is commonly used by the literature \cite{engel2017direct, Schops2019CVPR}. The estimated trajectory is first aligned with the ground truth by matching poses with the same timestamps. The RMSE of ATE is then computed by averaging the translational differences between the aligned trajectories. In addition, we also use the percentage of frame drops to measure the robustness of different algorithms.    

\boldparagraph{Motivating example} To clearly motivate the need for motion blur aware VO, we first evaluate the performance of ORB-SLAM and DSO on the sharp images and the corresponding motion blurred images respectively, on the ArchVizInterior dataset. See Section~\ref{sec:datasets} for details on the dataset. \tabnref{tab_archviz_dso_orb_result} demonstrates that motion blurred images affect both ORB-SLAM and DSO, in terms of estimated trajectory accuracy and robustness. Although there is no large accuracy drop for ORB-SLAM, there are significant frame drops. ArchVizInterior dataset collects images around a local area and scene overlaps exist among almost every image. Even though there are many frame drops, ORB-SLAM can still recover back due to its relocalization module, once the image is in better quality, assuming there is enough visual overlap with previously mapped areas. However, a reset might need to perform if the camera tranverses in un-explored scenes. There is no frame drops for DSO. However, its accuracy drops with a large margin compared to that with sharp images. 

Note that we rendered the dataset with the same trajectories from ETH3D \cite{Schops2019CVPR} dataset. The results demonstrate that the camera motion is not the main factor, which leads the failure of both ORB-SLAM and DSO on ETH3D dataset. It further justifies our motivation to create new datasets.

\begin{table*}
	\centering
	\small{
		\setlength\tabcolsep{4pt}
		\begin{tabular}{lccccccccccccc} \toprule
			& \multicolumn{6}{c}{ORB-SLAM~\cite{mur2017orb}} && \multicolumn{6}{c}{DSO~\cite{engel2017direct}} \\
			\cmidrule{2-7} \cmidrule{9-14} 
			& \multicolumn{2}{c}{ArchViz-1} & \multicolumn{2}{c}{ArchViz-2} & \multicolumn{2}{c}{ArchViz-3} && \multicolumn{2}{c}{ArchViz-1} & \multicolumn{2}{c}{ArchViz-2} & \multicolumn{2}{c}{ArchViz-3} \\
			\cmidrule{2-7} \cmidrule{9-14} 
			& ATE (m) & FD (\%) & ATE (m) & FD (\%) & ATE (m) & FD (\%) && ATE (m) & FD (\%) & ATE (m) & FD (\%) & ATE (m) & FD (\%) \\
			\midrule %\cmidrule{1-7} \cmidrule{9-14}
			Sharp & 0.0201 & 0 & 0.0048 & 0 & 0.0138 & 0 && 0.0196 & 0 & 0.0043 & 0 & 0.0140 & 0 \\ %\midrule
			Blur & 0.0325 & 22.1 & 0.0122 & 1.068 & 0.1008 & 19.5 && 0.2132 & 0 & 0.1655 & 0 & 0.1286 & 0  \\ 
			%\cmidrule{1-7} \cmidrule{9-14} 
			%
			Deblur & 0.0179 & 15.6 & 0.0066 & 2.763 & 0.0197 & 16.7 && 0.2065 & 0 & 0.1613 & 0 & 0.0481 & 0 \\ \bottomrule		    
		\end{tabular}
		\vspace{1ex}
	}
	\caption{\textit{The performance of both ORB-SLAM and DSO on the ArchVizInterior dataset}. \textit{Sharp}, \textit{Blur} and \textit{Deblur} denote the pipeline is running on the ground truth sharp images, motion blurred images and the deblurred images by DeblurGANv2 \cite{Kupyn2019ICCV} respectively. The FD column shows the percentage of dropped frames. Both the ATE and FD metrics are the smaller the better.} 
	\label{tab_archviz_dso_orb_result}
\end{table*}

\boldparagraph{Ablation studies} Since ArchvizInterior dataset has ground truth sharp images, which are paired with the motion blurred images, we conduct ablation studies with it for better comparisons. Our ablation studies consist of two parts, the selection of the deblurring network and experiments to demonstrate the effectiveness of our motion blur aware tracker. 

We evaluate the generalization performance as well as efficiency of both deblurring networks, \ie SRNDeblurNet \cite{Tao2018CVPR} and DeblurGANv2-mobileNet \cite{Kupyn2019ICCV} on the ArchvizInterior dataset. The evaluation is conducted with a laptop grade Nvidia RTX 2080 graphic card.
\begin{table}
	\small
	\begin{tabular}{lccc}
		\toprule
		   & PSNR (dB) $\uparrow$ & SSIM $\uparrow$ & Time (ms) \\
		\midrule
		Blur image & 26.80 & 0.7887 & N.A. \\
		DeblurGANv2m \cite{Kupyn2019ICCV} & 28.66 & 0.8156 & 38.1 \\
		SRNDeblurNet \cite{Tao2018CVPR} & 30.01 & 0.8491 & 140.3\\
		\bottomrule
	\end{tabular}
	\vspace{0ex}
    \caption{Generalization performance of DeblurGANv2-mobileNet \cite{Kupyn2019ICCV} and SRNDeblurNet \cite{Tao2018CVPR} on the ArchVizInterior dataset.}
    \label{tab_deblurNet}
\end{table}
\normalsize
\tabnref{tab_deblurNet} demonstrates that the DeblurGANv2-mobileNet is able to run in real time on a high-end GPU. However, its deblurring performance is worse than SRNDeblurNet as an expense. To verify if current performance of DeblurGANv2-mobileNet is sufficient to improve the performance of VO algorithms, we deblurred every image of ArchVizInterior dataset by DeblurGANv2-mobileNet. We run both ORB-SLAM and DSO with the deblurred images. The experimental results shown in \tabnref{tab_archviz_dso_orb_result} demonstrate that it can only improve the performance of VO algorithms with motion blurred images with a small margin. The reason is that the DeblurGANv2~\cite{Kupyn2019ICCV} has limited generalization performance as an expense for smaller model size. It demonstrates that the naive way to deblur every input frame (with an efficient deblurring network for real time operation), and feed the deblurred images to a standard VO pipeline is not the correct way to make VO robust to motion blur. It justifies our motivation to do hybrid motion blur aware VO, which can take advantage of a more powerful deblurring network with a larger model size. Our hybrid approach recovers the camera motion of severe blurred images by the motion blur aware direct image alignment algorithm, without the need to deblur them in frame rate. Since SRNDeblurNet takes around 140 ms to process a $752\times480$ pixels resolution image, which is sufficient to deblur selected key-frame image, and delivers better deblurring performance, we thus use it for our pipeline.

To study the effectiveness of MBA-VO, we experiment with sharp images and blurry images respectively. Experimental results from \tabnref{tab_archviz_our_result} demonstrate that MBA-VO is able to achieve similar performance as both ORB-SLAM and DSO if the images are not motion blurred. For motion blurred images, MBA-VO is able to achieve competitive accuracy as that for sharp images without any frame drops. 
To further demonstrate the effectiveness of our motion blur aware tracker, we set the camera exposure time to be 0 during the estimation of the camera poses (\ie~\eqnref{eq_photoconsistency2}). It enforces the tracker to assume the current blurry images as sharp images and do normal pose estimations instead. The other settings are kept the same (\eg we still use SRNDeblurNet to deblur the keyframe images). The resulted ATE metrics are 0.22 m, 0.1558 m and 0.2113 m respectively for the ArchVizInterior dataset. The experimental results thus demonstrate the necessity to do motion blur aware tracking. 

\begin{figure*}
	\small{
	\begin{tabular}{ccc}
		\includegraphics[width=0.305\textwidth]{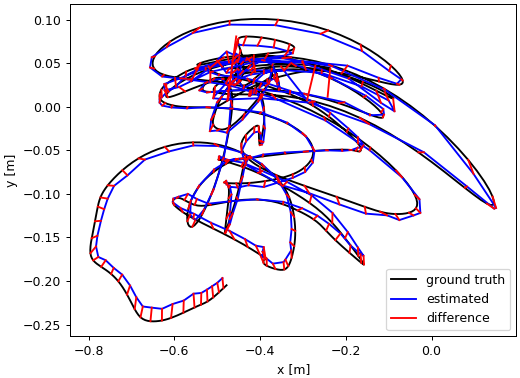} &
		\includegraphics[width=0.295\textwidth]{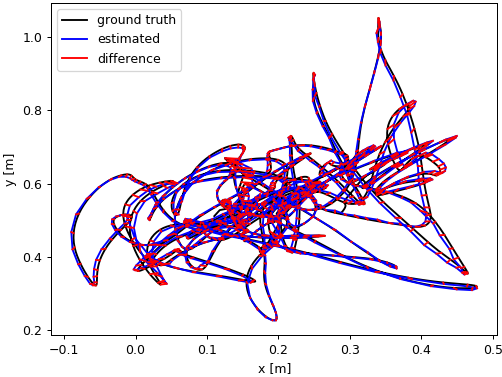} & 
		\includegraphics[width=0.30\textwidth]{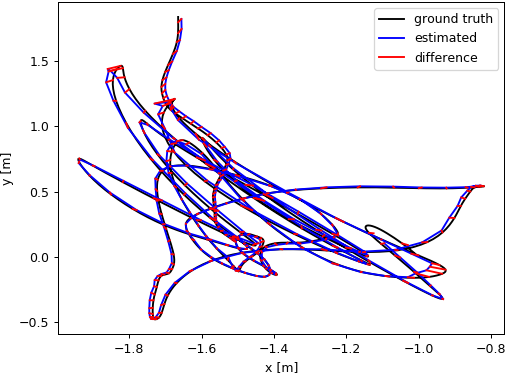} \\
		ArchViz-1 & ArchViz-2 & ArchViz-3
	\end{tabular}
    }
    \caption{Estimated trajectories of MBA-VO from the motion blurred image sequences of the ArchVizInterior dataset. It demonstrates that MBA-VO can estimate accurate trajectories, although the camera motions are very challenging.}
    \label{fig:mba_vo_traj_archviz}
\end{figure*}

\figref{fig:mba_vo_traj_archviz} demonstrates the estimated trajectories of MBA-VO on the motion blurred sequences from ArchVizInterior dataset. 
Both the quantitative and qualitative results demonstrate the effectiveness of our proposed algorithm for motion blurred image sequences. 

\begin{table}
	\centering
	\small{
		\setlength\tabcolsep{2.5pt}
		\begin{tabular}{lcccccc} \toprule
			& \multicolumn{2}{c}{ArchViz-1} & \multicolumn{2}{c}{ArchViz-2} & \multicolumn{2}{c}{ArchViz-3} \\
			\cmidrule{2-7} 
			& ATE (m) & FD (\%) & ATE (m) & FD (\%) & ATE (m) & FD (\%)  \\ \midrule
			%\cmidrule{1-7}
			%
			Sharp & 0.0197 & 0 & 0.0101 & 0 & 0.0157 & 0 \\ 
			Blur & 0.0256 & 0 & 0.0184 & 0 & 0.0202 & 0  \\ 
			\bottomrule		    
		\end{tabular}
		\vspace{1ex}
	}
	\caption{\textit{The performance of MBA-VO on the ArchVizInterior dataset}. \textit{Sharp} and \textit{Blur} denote the pipeline is running on the ground truth sharp images and motion blurred images respectively. FD is the percentage of dropped frames.} 
	\label{tab_archviz_our_result}
\end{table}

\boldparagraph{Evaluation with TUM RGB-D dataset}
To evaluate the performance of MBA-VO with real motion blurred images, we select three sequences with large motion blur from the TUM RGB-D dataset \cite{Sturm2012IROS}. In particular, we select the \textit{fr1-desk}, \textit{fr1-desk2} and \textit{fr1-room} from the handheld SLAM category. Since the camera is hand held, which is similar to head-mounted camera, hand shaken would result in fast rotational motion though the translational velocity is small. For augmented/virtual/mixed reality applications, head rotational motion is the main cause of severe motion blur. 

\begin{table}
	\centering
	\small{
		\setlength\tabcolsep{1.5pt}
		\begin{tabular}{lcccccc} \toprule
			& \multicolumn{2}{c}{fr1-desk} & \multicolumn{2}{c}{fr1-desk2} & \multicolumn{2}{c}{fr1-room} \\
			\cmidrule{2-7} 
			&ATE (m) & FD (\%) & ATE (m) & FD (\%) & ATE (m) & FD (\%)  \\ \midrule
			%\cmidrule{1-7}
			%
			\small{ORBSLAM} & 0.1781 & 5.1 & 0.3005 & 33.8 & 0.0657 & 46.5\\ 
			DSO & 0.4956 & 0 & 0.7762 & 0 & 0.2992 & 0 \\ 
			%\cmidrule{1-7} 
			%
			MBA-VO & 0.1021 & 0 & 0.3997 & 0 & 0.1435 & 0 \\ 
			\bottomrule		    
		\end{tabular}
	}
    \vspace{0ex}
	\caption{\textit{Comparison on TUM RGB-D dataset~\cite{Sturm2012IROS}}. ORB-SLAM suffers from significant frame drops, although it provides accurate estimates. The proposed method, MBA-VO, improves on DSO and provides more accurate estimates with no frame drops.} 
	\label{tab_archviz_our_tum_result}
	\normalsize
\end{table}

\tabnref{tab_archviz_our_tum_result} demonstrates the performance of MBA-VO against ORB-SLAM and DSO on sequences with large motion blur from TUM RGB-D dataset. It demonstrates that MBA-VO is able to improve the accuracy over the original DSO algorithm, while is also more robust compared to sparse feature based approach, with motion blurred images. Note that ORB-SLAM suffers from significant frame-drops, although it generally provides more accurate poses for these sequences (low ATE).

\boldparagraph{Evaluation with our real-world dataset} Since the goal of the TUM RGB-D dataset is not evaluating the robustness of monocular VO/SLAM algorithms, we created a specific large real dataset with varying levels of motion blur (see Section~\ref{sec:datasets}). We evaluate ORB-SLAM, DSO and MBA-VO with it. Due to space limit, we present the experimental results from a subset of the datasets in \tabnref{tab_archviz_MBADSO_our_result}. More experimental results can be found from our supplementary material. The experimental results illustrate that ORB-SLAM has significant frame drops in general for all the sequences, while it usually is more accurate. Compared to DSO, which has no frame drops, MBA-VO has better accuracy. In general, MBA-VO achieves better robustness and accuracy, compared to both DSO and ORB-SLAM on the real motion blurred image sequences.

\begin{table}
	\centering
	\small{
		\setlength\tabcolsep{3pt}
		\begin{tabular}{lcccccc} \toprule
			& \multicolumn{2}{c}{ORB-SLAM~\cite{mur2017orb}} & \multicolumn{2}{c}{DSO~\cite{engel2017direct}} & \multicolumn{2}{c}{MBA-VO} \\
			\cmidrule{2-7} 
			& ATE (m) & FD (\%) & ATE (m) & FD (\%) & ATE (m) & FD (\%)  \\\midrule
			Seq0 & 0.1265 & 7.0 & 0.2722 & 0 & \bf 0.0581 & 0 \\ 
			Seq1 & 0.0839 & 36.8 & 0.4327 & 0 & \bf 0.0692 & 0 \\ 
			Seq2 & 0.1992 & 11.9 & 0.1958 & 0 & \bf 0.0446 & 0 \\ 
			Seq3 & x & x & 0.4044 & 0 & \bf 0.1615 & 0 \\ 
			Seq4 & x & x & x & x & \bf 0.1323 & 0 \\ 
			\bottomrule		    
		\end{tabular}
	}
	\caption{\textit{The performance of MBA-VO on our dataset}. x denotes the corresponding algorithm fails on that particular sequence. It demonstrates that MBA-VO improves the accuracy of DSO, while being robust to motion blur with no frame drops.} 
	\label{tab_archviz_MBADSO_our_result}
\end{table}

\boldparagraph{Discussions} The experimental results demonstrate that DSO~\cite{engel2017direct} generally performs worse than ORB-SLAM~\cite{mur2017orb} for motion blurred images, in terms of the ATE metric. It is caused by the datasets we evaluated on have many more severely blurred images. In this case, ORB-SLAM~\cite{mur2017orb} simply discards the severe blurred images without affecting the accuracy of the remaining frames. In contrast, DSO~\cite{engel2017direct} does not drop the frames, leading to the overall loss of accuracy due to including more challenging frames in the estimation. Note that the ATE metric is only computed from the successfully tracked frames.

\section{Conclusion}
We present a hybrid visual odometry algorithm which is robust to motion blur. The camera motion trajectory within exposure time is explictly modelled and estimated during tracking. It allows us to compensate the effect of motion blur without deblurring all of them. We also propose a novel benchmarking dataset targeting motion blur aware visual odometry. Experimental results demonstrate that our algorithm improves the accuracy and robustness over existing methods on both synthetic and real-world datasets. While we only consider monocular VO, our approach could also be applied to other settings such as VIO and RGB-D SLAM. We believe both our method and dataset would be a valuable step towards the era of robust visual odometry.

{\small
\bibliographystyle{ieee_fullname}
\bibliography{ms}
}

\end{document}